\documentclass[letterpaper,12pt]{wlscirep}
\usepackage{amsmath}
\usepackage{amssymb}
\usepackage{multirow}
\usepackage{booktabs}
\usepackage{amsthm}
\usepackage{tikz}
\usepackage{xcolor}
\usepackage{url}

\usepackage{pgf}
\usetikzlibrary{shapes,arrows,automata}

\definecolor{RyansColor}{rgb}{0,0.5,0}

\title{Whetstone: A Method for Training Deep Artificial Neural Networks for Binary Communication}

\author[1,*]{William~Severa}
\author[1]{Craig~M.~Vineyard}
\author[1]{Ryan~Dellana}
\author[1]{Stephen~J.~Verzi}
\author[1,*]{James~B.~Aimone}
\affil[1]{Center for Computing Research, Sandia National Laboratories, Albuquerque, NM, 87185 USA  \{wmsever, cmviney, rdellan, sjverzi, jbaimon\}@sandia.gov}
\affil[*]{Correspondence should be addressed to W.S. or J.B.A.}
		
\begin{abstract}
Deep artificial neural networks are becoming pervasive in computing domains, particularly as their impact spreads beyond basic image classification.  However, the computational cost of deep networks, both in training and inference modes, presents challenges to deploying these algorithms broadly. Research into reduced precision networks and compressing trained networks seeks to reduce the computational cost of using deep learning, however streamlining the communication within deep learning systems is also widely impactful. 

This paper presents a new technique for training networks for low-precision communication.  Targeting minimal communication between nodes not only enables the use of emerging spiking neuromorphic platforms, but may additionally streamline processing conventionally.  Low-power and embedded neuromorphic processors potentially offer dramatic performance-per-Watt improvements over traditional von Neumann processors, however programming these brain-inspired platforms generally requires platform-specific expertise which limits  their applicability.  To date, the majority of artificial neural networks have not operated using discrete spike-like communication.  

  We present a method for training deep spiking neural networks using an iterative modification of the backpropagation optimization algorithm.  This method, which we call Whetstone, effectively and reliably configures a network for a spiking hardware target with little, if any, loss in performance.  Whetstone networks use single time step binary communication and do not require a rate code or other spike-based coding scheme, thus producing networks comparable in timing and size to conventional ANNs, albeit with binarized communication.  We demonstrate Whetstone on a number of image classification networks, describing how the sharpening process interacts with different training optimizers and changes the distribution of activity within the network.  We further note that Whetstone is compatible with several non-classification neural network applications, such as autoencoders and semantic segmentation.  Whetstone is widely extendable and currently implemented using custom activation functions within the Keras wrapper to the popular TensorFlow machine learning framework.
\end{abstract}

\begin{document}

\maketitle
\section{Introduction}
\subsection{Overview and Motivation}
Artificial neural network (ANN) algorithms, specifically deep convolutional networks (DCNs) and other deep learning methods, have become the state-of-the-art techniques for a number of machine learning applications~\cite{he2016deep, pinheiro2015learning, lecun2015deep}.  While deep learning models can be expensive both in time and energy to operate and even more expensive to train, their exceptional accuracy on fundamental analytics tasks such as image classification and audio processing has made their use essential in many domains.

Some applications can rely on remote servers to perform deep learning calculations; however, for many applications such as onboard processing in autonomous platforms like self-driving cars, drones, and smart phones, the resource requirements of running large ANNs may still prove to be prohibitive ~\cite{cvpr_2017_yang_energy, coppola2017driverless}. Large ANNs with many parameters require a significant storage capacity which is not always available. And data movement energy costs are greater than that of performing computation making large ANNs intractable ~\cite{horowitz20141}. Additionally onboard processing capabilities are often limited to meet energy budget requirements further complicating the challenge. Other factors such as privacy and data sharing also provide a motivation for performing computation locally rather than on a remote server.

The development of specialized hardware to enable more efficient ANN calculations seeks to facilitate moving ANNs into resource-constrained environments, particularly for trained algorithms that simply require the deployment of an inference-ready network. The current generation of specialized ANN processors gain advantages simply by customizing key computational kernels for ANNs in application-specific integrated circuits (ASICs). Examples include the Google Tensor Processing Unit (TPU), the Intel Nervana Neural Network Processor, and the NVIDIA Volta GPU architecture with dedicated tensor cores ~\cite{jouppi2017datacenter, rao_2018, hemsoth_2018,  markidis2018nvidia}. Many of these emerging accelerators targeting deep neural network inference enable impressive numbers of operations such as multiply and accumulates (MACs), however these measures by themselves do not fully capture the emerging landscape of neural network architectures. For instance, it has been shown that the number of MACs is not indicative of the energy consumption of a DCN ~\cite{chen2018understanding,yang2016designing}, and so hardware supporting high operation counts may require a higher power budgets than some use cases can afford. 

Rather than optimizing a key computational kernel, further benefits will likely be realized by hardware platforms that implement versions of ANNs that are mathematically optimized for efficient performance, such as using low-precision weights (minimize memory costs) and discrete activation functions (minimize communication costs). One such promising technology is neuromorphic hardware, which has been shown to be capable of running ANNs and can potentially offer orders-of-magnitude lower power consumption (i.e., performance-per-Watt) than more conventional digital accelerators~\cite{james2017historical}.  While there are a number of approaches to neuromorphic hardware, most scalable current platforms such as SpiNNaker and IBM TrueNorth achieve low-power performance by coupling spike-based data representations with brain-inspired communication architectures~\cite{merolla2014million, khan2008spinnaker, schuman2017survey, james2017historical, knight2016large, sze2017efficient}.  

For spiking neuromorphic hardware to be useful, however, it is necessary to convert an ANN, for which communication between artificial neurons can be high-precision, to a spiking neural network (SNN). \textit{Spiking }is a generic term describing the abstraction of action potential formation in biological neurons crossing a voltage threshold. What ``spiking'' means for SNNs varies considerably, but at minimum it requires that neurons only communicate a discrete event (a `1') or nothing. In addition, many approaches to SNNs leverage time --- or when the spike occurs --- as a key state variable to communicate information; although the use of time varies considerably across approaches. In this paper, which focuses on the first step of mapping SNNs to static ANN applications like image classification, we define `spiking' as the simplest activation function for neurons that is compatible with neuromorphic hardware --- a discrete $1$ or $0$ threshold activation. For static ANN applications, this time-agnostic approach is sufficient (and in fact desirable for reasons of throughput), however for dynamic applications such as video processing, it is likely that the time dimension offered by spiking will be useful, though that is beyond the scope of this study.

The conversion of ANNs to SNNs---whatever their form---is non-trivial, as ANNs depend on gradient-based backpropogation training algorithms, which require high-precision communication, and the resultant networks effectively assume the persistence of that precision.  While there are methods for converting existing ANNs to SNNs, these transformations often require using representations that diminish the benefits of spiking. 
Here, we describe a new approach to training SNNs, wherein the ANN training is to not only learn the task, but to produce a SNN in the process.  Specifically, if the training procedure can include the eventual objective of low-precision communication between nodes, the training process of a SNN can be nearly as effective as a comparable ANN.  This method, which we term \textit{Whetstone} (portrayed in Fig.~\ref{WhetstoneMethod}) inspired by the tool to sharpen a dull knife, is intentionally agnostic to both the type of ANN being trained and the targeted neuromorphic hardware.  Rather, the intent is to provide a straightforward interface for machine learning researchers to leverage the powerful capabilities of low-power neuromorphic hardware on a wide range of deep learning applications (see Methods \ref{Methods_Implementation}).

\section{Results}
\label{results}

\subsection{Whetstone Method Converts General ANNs to Spiking NNs}

\begin{figure}[!ht]
\includegraphics[width=5.5in]{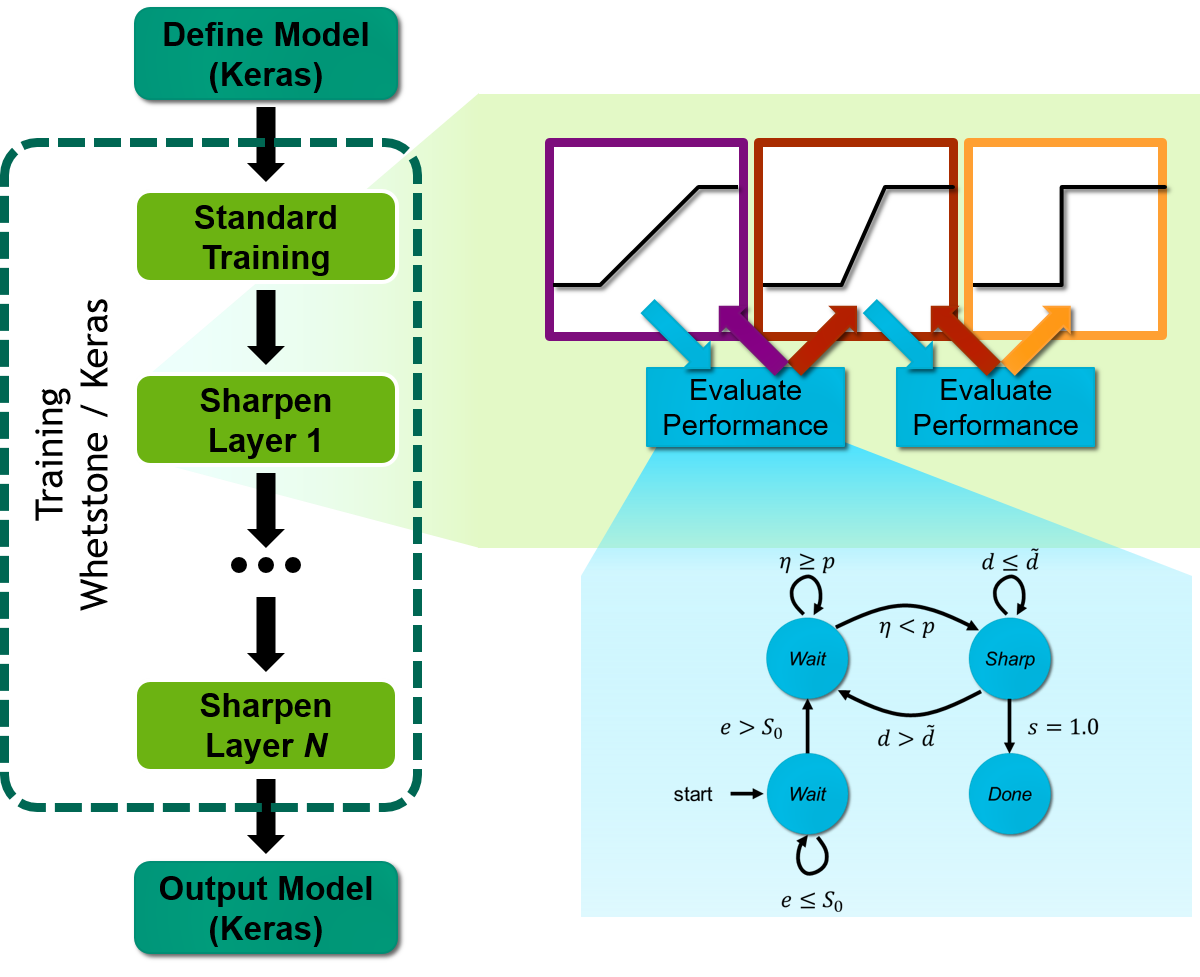}
\caption{\textbf{Overview of Whetstone Process.} Whetstone is a process for training binary, threshold-activation spiking neural networks using existing deep learning methods.  By adjusting neuron activation functions \textit{during training} (green inset), the network more and more closely approximates the behavior a spiking neural network.  The sharpening process is  automated using an adaptive sharpening schedule (blue inset).  In contrast to rate-coded or temporally-coded spiking neurons, the neurons here are instantaneously acting in that each input is evaluated over a single timestep, an approach compatible with both deep learning frameworks and many neuromorphic processors.  The final trained network is then portable and can be easily instantiated on neurormorphic and conventional platforms.}
\label{WhetstoneMethod}
\end{figure}

The Whetstone algorithm operates by incorporating the conversion into binary activations directly into the training process.  Because most techniques to train ANNs rely on stochastic gradient descent methods, it is necessary that the activations of neurons be differentiable during the training process.  However, as networks become trained, the training process is able to incorporate additional constraints, such as targeting discrete communication between nodes.  With this shift of the optimization target in mind, Whetstone gradually pushes the network towards discrete ``spike'' activations by shifting the gradient of activation functions incrementally towards a discrete perceptron-like step function, then fine-tuning the network to account for any loss as a result of that conversion (see Methods \ref{Methods_SharpeningSection}).  By gradually ``sharpening'' neurons' activations layer-by-layer, the network can slowly approach a spiking network that has minimal loss from the full-precision case.  The Whetstone process is explained in full detail within the Methods section.

\begin{figure}[!ht]
\includegraphics[width=7in]{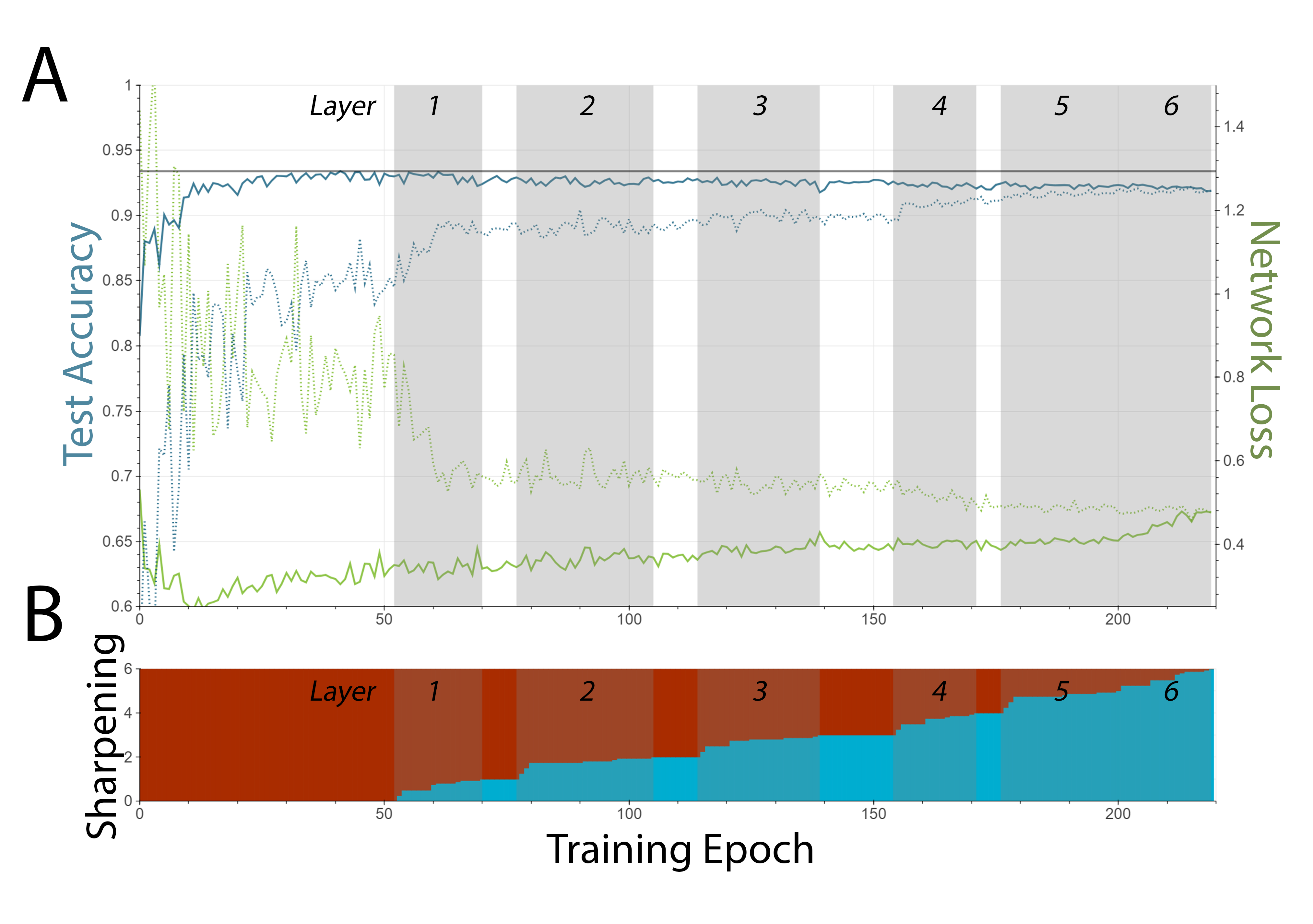}
\caption{\textbf{Training of Single Network through Whetstone Process.}  (A) A six-layer ANN was trained on the MNIST data set, reaching a steady-state accuracy (black line) around 50 epochs.  Within the shaded regions, the Whetstone process is gradually pushing neurons, layer-by-layer, to have progressively sharper activation functions until they are essentially discrete. The dotted line shows the accuracy (or loss) of the network if the network is forced to fully discrete activations at that time; whereas the solid line shows the accuracy (or loss) of the Whetstone network.  The observed test accuracy drop is due to the precision drop of communication between layers. (B) The sharpening of layers of the network shown in panel A.  Each shaded region shows the relative sharpening of the layer over a number of epochs.  A layer that is fully blue during that sharpening phase is discrete, and the height of blue within each shaded region reflects how sharp that layer is through the sharpening phase.}
\label{OverviewWhetstone}
\end{figure}

The outputs of Whetstone are shown in Fig.~\ref{OverviewWhetstone} for the training of an exemplar network.  The goal of Whetstone training is to produce an ANN with discrete activations (either $1$ or $0$) for all communication between neurons.  However, because networks are not typically trained with this goal incorporated into their optimization, the immediate conversion of activations into a binary $1$ or $0$ results in a substantial drop in accuracy.  However, as Whetstone gradually converts networks to spiking through the incremental sharpening of each layer one-by-one (Fig.~\ref{OverviewWhetstone}B), the performance of the Whetstone networks only experiences minor impairment compared to the standard trained network.  Furthermore, once the early layers are discretized through Whetstone, the loss introduced by forcing networks to have discrete communication is minimized.

\subsection{Description of Baseline Spiking Accuracy}

\begin{figure}[!ht]
\includegraphics[width=7in]{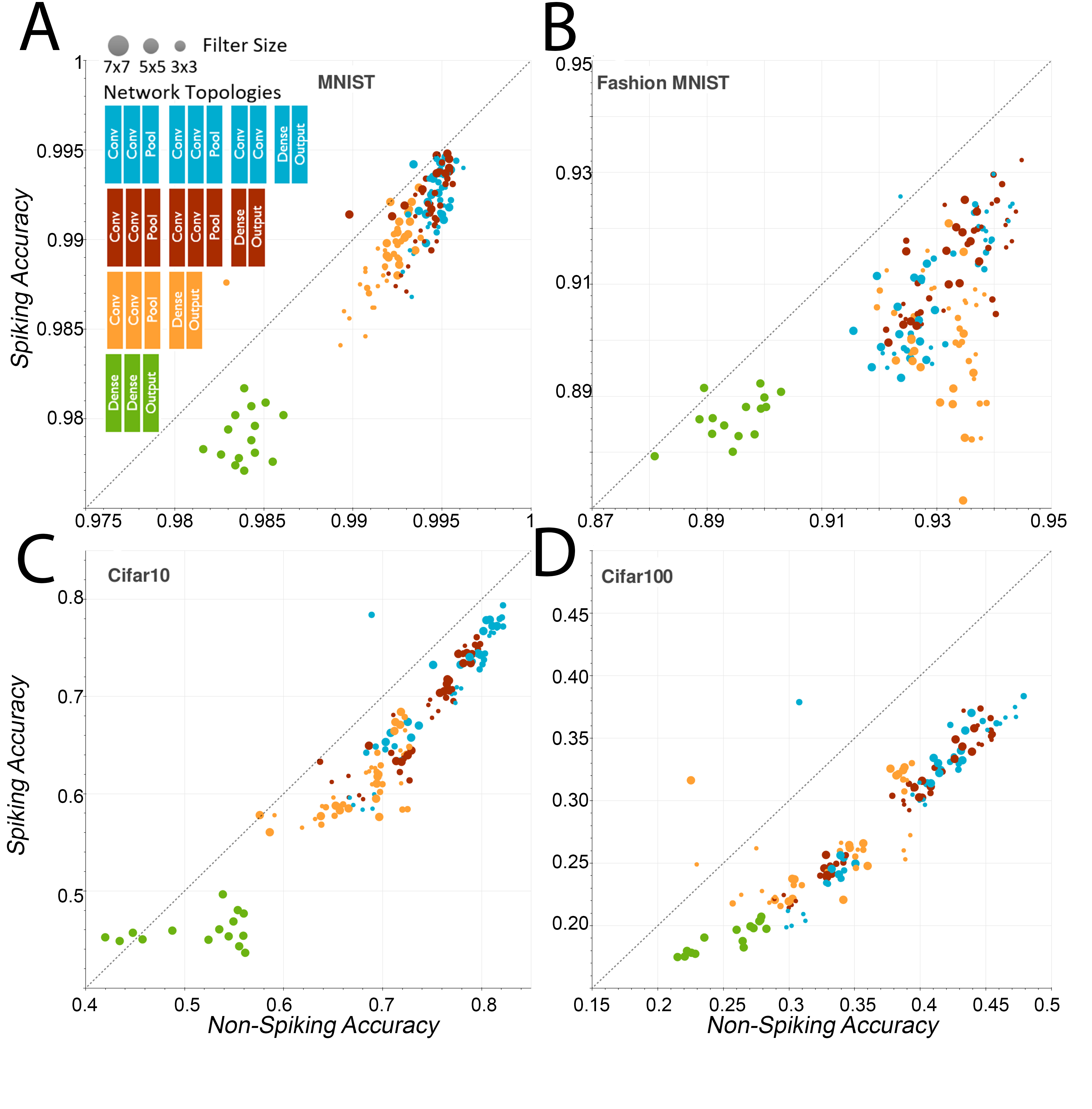}
\caption{\textbf{Various network topologies and datasets with their original accuracy and corresponding spiking accuracy.}  Whetstone was tested on several network sizes and topologies against the MNIST, Fashion MNIST, CIFAR-10, and CIFAR-100 data sets.  The distance below (or above) the diagonal indicates the penalty (or improvement) that the binarized network exhibited.  Not surprisingly, larger networks (yellow, red, blue) had higher classification accuracy than the smallest network (green) for both spiking and non-spiking networks.  For most cases, spiking accuracy was moderately lower than the non-spiking accuracy, indicating a small penalty in classification accuracy due to the reduced precision.}
\label{performance}
\end{figure}

We examined the performance of Whetstone on image classification within four different data sets: MNIST, Fashion MNIST, CIFAR-10, and CIFAR-100, and several different network configurations.  We plot the performance of networks across a wide hyperparameter space in Fig.~\ref{performance}.  In general, hyperparameter optimization for deep neural networks is a complex, open problem with several popular solutions~\cite{bergstra2013hyperopt, li2017hyperband}.  These methods are generally compatible with our approach as they do not depend on the specific activation functions, and so in a production environment the hyperparameters of Whetstone networks can be optimized by industry-standard approaches. We hope that, in providing this wide scope of networks and performance levels, we can gain insight into Whetstone's performance across applications and hyperparameters rather than only present the hand-tuned top-performing networks in Table~\ref{table:comparisonTable}.  For most experiments, equivalent spiking networks were somewhat more brittle leading to modest overall performance losses, as shown in Fig.~\ref{performance}.  This is not surprising, given that the spike representations means less precision in the communication between layers, and the relatively small differences suggest that small specializations to networks for spiking may mitigate much of this loss.

Within the configurations tested, there was not a common trend to suggest a coarse consideration would improve spiking performance.  For instance, deeper networks leveraging convolutional layers performed better for both non-spiking networks and their spiking equivalents, as one would expect.  For MNIST, the largest network had roughly equivalent spiking and non-spiking performance, however this is not the case for other data sets.  Likewise, we observed some runs where larger kernel sizes were helpful for the spiking networks, however this was not universally the case.

While the modest penalty for spiking that we observed may be permissible for some applications where the energy savings of a spiking representation would outweigh the accuracy hit, we sought to further improve Whetstone performance by examining three aspects of network training that may uniquely impact the spike-conversion process.  First, we examined strategies for output encoding as we observed that a number of spiking runs occasionally suffered from whole classes failing to be classified.  Second, we observed that the choice of optimizer and associated learning rates is non-trivial given the changes to network encodings over time contributed by the Whetstone process.  Finally, we implemented batch normalization, which is not obviously consistent with a streaming spiking representation, but nonetheless can improve performance.

\subsection{$\mathbf{N}$-Hot output encoding and addressing `dead' nodes}\label{EncodingSection}
One challenge of the bounded rectified linear units (bRELUs) used in Whetstone is that nodes may stop responding to \textit{any} inputs, effectively rendering them `dead' nodes.  This is particularly an issue at the output layer, where if a node ceases to respond a class may no longer be represented at all.  These encoding failures have been noted for conventional networks, both utilizing sigmoids and RELU, particularly in the context of transfer learning or other applications where a subset of classes cease to be trainable.  However, because our sharpening process can move a node's encoding from a differentiable to a non-differentiable regime, it is likely that the problem is exacerbated here.  

\begin{figure}
\includegraphics[width=6.5in]{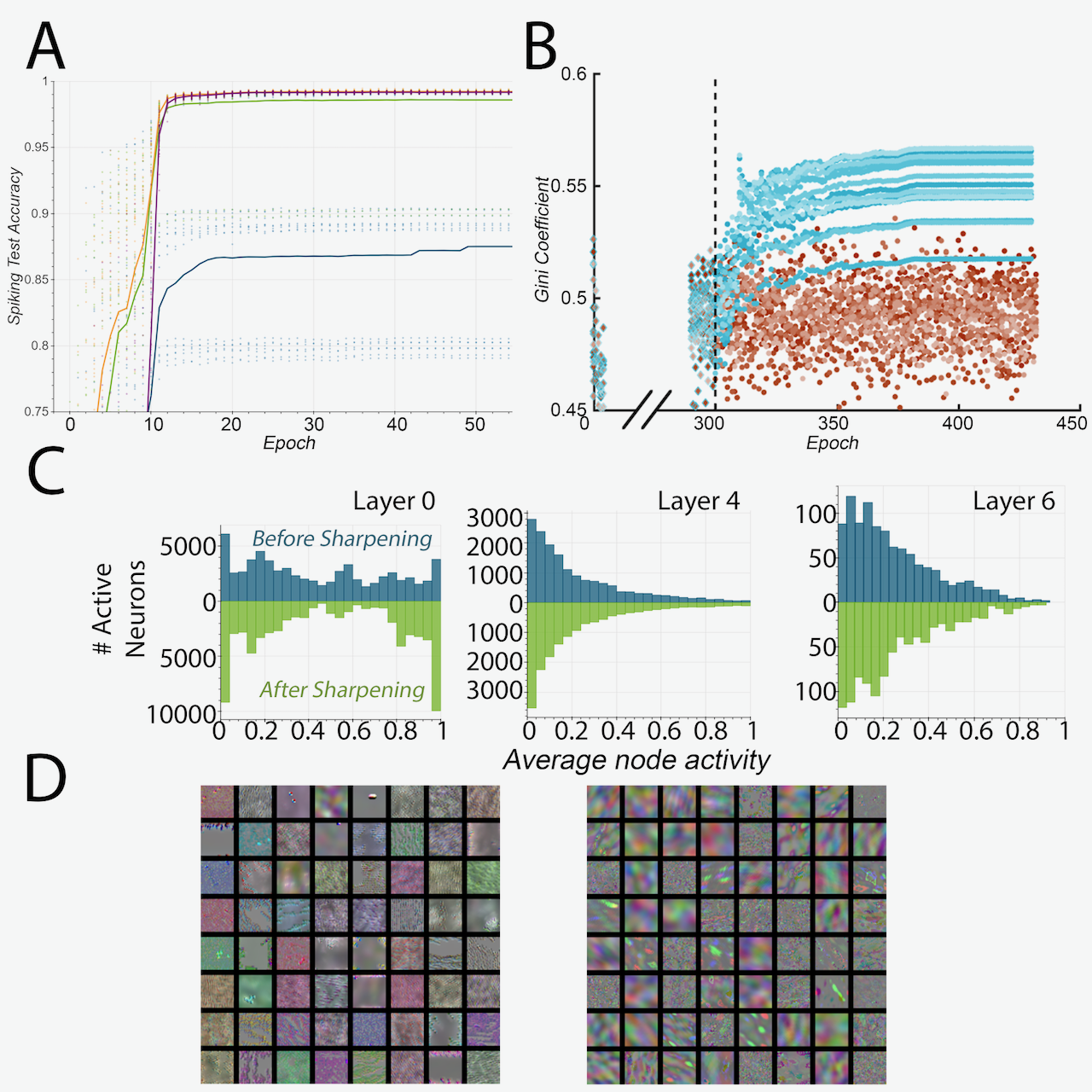}
\caption{\textbf{Whetstone training skews encoding to fewer nodes, requiring $N$-hot output encodings.} (A) As shown by the blue line, $1$-hot encoding of outputs often fails due to dead nodes.  $2$-hot encodings (green) are more stable, but do occasionally suffer from impaired performance due to dead nodes.  $4$-hot (yellow) and $8$-hot (red) encodings are more stable.  (B) Sharpening of neuron activations increases Gini coefficient of trained networks.  During conventional training (Red), there is a moderate increase of Gini, suggesting a moderate inequality of neuron use in encoding training data; however once Whetstone (Blue) begins to sharpen activation functions, starting with input layers, the activity of neurons becomes more concentrated within a smaller population of neurons, as indicated by the reliable increase in Gini coefficients. (C) Distribution of average node activity before (above x-axis) and after (below x-axis) Whetstone sharpening.  Layer $0$ (left) shows a rather broad activation of neurons initially, yet after Whetstone most neurons are either active most of the time or never activated.  Deeper layers are more sparsely active, yet the distribution does not change as much during sharpening. (D) Activation maximizations of first-layer filters in non-spiking (left) and spiking (right) CIFAR classification networks.}
\label{output_encodings}
\end{figure}

The results reported in Fig.~\ref{performance} used a conservative encoding scheme to avoid any loss due to `dead' nodes in the output layer, in which each output class is represented by redundant neurons that independently determine if that class is activated (see Methods \ref{Methods_Encodings}).  To examine what sizes of output layers are adequate, we tested networks with $1$-, $2$-, $4$-, $8$-, and $16$-hot encoding.  Immediately noticeable is that, as predicted from aforementioned observation of dead nodes, $1$-hot encoding is unreliable and insufficient for our purposes.  Any classes that were represented by a dead node introduced misclassification of that entire class, which equated to$~10\%$ penalties on MNIST (Fig.~\ref{output_encodings}A).  

Incorporating more output nodes, implementing $2$-hot and $4$-hot (Fig.~\ref{output_encodings}) encoding alleviated this problem.  The $2$-hot encoding example shows that a single `living' node is sufficient to achieve high accuracy, however there remains a small but non-trivial possibility that both output nodes for a class may die.  We did observe one such case out of $30$ runs where a network had a test classification accuracy of $~90\%$ indicating that one class was dead.  $4$-hot encoding does not suffer from the same problem.  We never observed a case where a class's entire complement of output nodes all died, though it is possible that with enough classes this may eventually occur by chance.  Nonetheless, the $4$-hot encoding appeared to offer an advantage beyond just size, with that architecture showing high overall performance on MNIST ($99.24\%$), in this study surpassing the performance of equivalent networks with larger output layers ($8$-hot and $16$-hot).  

This observation suggests that if the effect of dead nodes could be mitigated directly instead of by using $N$-hot encoding, network performance may be improved further.  To avoid the dead nodes in the output layer, we then attempted to replace the bRELU, which we know is susceptible to death upon sharpening, with a sigmoid layer followed by a softmax function (data not shown).  The network performance is quite strong with only one output neuron per class; however, the final step which discretizes the output to a spiking form causes a sharp degradation.  This non-graceful degradation of performance upon conversion to spiking supports the choice of bRELUs for the initial network training.  Nevertheless, this result suggests that if output bRELUs could be kept alive, considerably higher network performance could be achieved. 

As an initial exploration into the cause of this concentration, we measured the Gini Coefficient of all the neurons within a network as Whetstone sharpened it (Fig.~\ref{output_encodings}B).  The Gini Coefficient ranges from $0$ to $1$ and is commonly used as a metric of inequality in economics, and in the context of neural networks --- particularly those with bounded activation functions such as the bRELU used here --- we can use it to measure the relative efficiency of a coding scheme across the neurons within a network.  A high Gini would be indicative of a small subset of neurons being used to encode most information, whereas a low Gini would indicate that the full population of neurons is used equivalently across all information.

As seen in Fig.~\ref{output_encodings}B, the sharpening of activations reliably increased the Gini Coefficient of our networks by roughly $10\%$.  This is consistent with the above finding that the sharpening of networks leads to dead nodes which effectively stop being used in the network.  Interestingly, the sharpening process has the greatest effect on the distribution of nodes during the sharpening of the early layers, as shown in the rapid rise in Gini at epoch $300$ (Fig.~\ref{output_encodings}B) as well as when the change in activity distribution before and after sharpening.  As shown in Fig.~\ref{output_encodings}C, in the first layer (Layer $0$), there is a much greater skew of values to either always be active or inactive after sharpening; whereas the distribution of intermediate layers average activations do not change considerably through the Whetstone process. 

Finally, we looked at how sharpening changes what inputs neurons respond to preferentially (Fig.~\ref{output_encodings}D).  Using the keras-vis package \cite{raghakotkerasvis}, we displayed what maximally activated each of the first layer filters in a non-spiking and spiking network trained on CIFAR.  While these results are qualitative, they do suggest that the sharpening process changes the distribution of spatial and spectral frequencies of the first layer convolutional filters in ways that may have implications on how information is distributed through the downstream layers of the network.

\subsection{Effects of reduced weight precision}

Another consideration for Whetstone networks is how they may be affected by the limited weight precision often seen in neuromorphic hardware.  While the representation of weights differs considerably across spiking neuromorphic platforms, these architectures typically do not use the full-precision floating-point representations available on conventional GPUs and CPUs. Therefore, we tested whether the reduced communication precision targeted by Whetstone is particularly vulnerable to reducing the weight precision as well. 

We tested the impact of reducing precision on Whetstone-spiking and non-spiking versions of dense and convolutional networks.  While methods exist to adjust the training of reduced precision weights, we limited this study to post-hoc precision reduction from full-precision (float32) networks as a worst-case baseline.  As seen in Table~\ref{precision_table}, for both dense and convolution networks, the conversion of 32-bit floating-point to fixed-point representations of Q4.16 (4 bits before the decimal including the sign bit, and 16 bits after the decimal) and Q4.8 had little effect on MNIST classification.  For dense networks, further decreases of precision to Q4.7, Q4.6, and Q4.5 had similar increases in the error rate for both spiking and non-spiking networks.  For convolutional networks, the increase in error rate due to low precision was similar for Q4.7 and Q4.6 fixed-point representations, but spiking networks were uniquely impacted by going down to Q4.5.

Overall, these results indicate that Whetstone networks are not uniquely affected by fixed-point representations at the levels of weight precision common to digital spiking neuromorphic hardware; however at very low precision representations it may be important to factor in the lower precision into the training process.  It is likely the case that such interventions will be necessary for both Whetstone and conventional network implementations.

\begin{table}
\caption{Accuracy of sharpened and non-sharpened networks at reduced precision.  Presented are the mean and range of accuracies for MNIST across ten sample networks each of two types.  Dense networks had two hidden layers ($512$ neurons each) and a $10$-hot output encoding. A small convolution network was chosen to give realistic, but conservative estimates of degradation.  The topology consists of two Convolution-MaxPool blocks and three dense layers before a $10$-hot output layer.  }
\label{precision_table}
\begin{tabular}{cc@{\hskip 0.5in}cccc}
\toprule
& &\multicolumn{2}{c}{Spiking} & \multicolumn{2}{c}{Non-Spiking}\\
& Precision & Mean & Range  & Mean & Range \\
\midrule
\multirow{6}{*}{Dense}&  float32&$0.9794$&$[0.9784,0.9820]$&$0.9854$&$[0.9837,0.9865]$\\
&Q4.16&$0.9794$&$[0.9777,0.9821]$&$0.9854$&$[0.9838,0.9865]$\\
&Q4.8&$0.9786$&$[0.9772,0.9803]$&$0.9849$&$[0.9836,0.9866]$\\
&Q4.7&$0.9773$&$[0.9757,0.9800]$&$0.9842$&$[0.9834,0.9855]$\\
&Q4.6&$0.9712$&$[0.9673,0.9742]$&$0.9798$&$[0.9774,0.9827]$\\
&Q4.5&$0.8679$&$[0.7732,0.9207]$&$0.8922$&$[0.8385,0.9447]$\\
\midrule
\multirow{6}{*}{Convolution}& float32 & $0.9815$ & $[0.9791,0.9836]$  & $0.9905$ & $[0.9896,0.9914]$\\
& Q4.16 & $0.9815$ & $[0.9789, 0.9835]$ & $0.9905$ & $[0.9896, 0.9914]$\\
& Q4.8 & $0.9815$ & $[0.9797, 0.9838]$  & $0.9905$ & $[0.9897, 0.9915]$\\
& Q4.7 & $0.9802$ & $[0.9782, 0.9817]$  & $0.9902$ & $[0.9894, 0.9916]$\\
& Q4.6 & $0.9754$ & $[0.9714, 0.9795]$  & $0.9884$ & $[0.9871, 0.9899]$\\
& Q4.5 & $0.9306$ & $[0.8867, 0.9482]$  & $0.9752$ & $[0.9639, 0.9813]$\\
\bottomrule
\end{tabular}
\end{table}

\subsection{Training optimizers and batch normalization impact Whetstone efficacy}

We next examined several different training optimizers on Whetstone networks. In general, the best optimization approach often differs considerably based on network architecture, training data, and initialization~\cite{ruder2016overview}.  As these different optimization techniques shape aspects of the stochastic gradient descent process to better navigate the energy landscape, it stands to reason that Whetstone, which is changing the gradients through continually sharpening the activation function, may interact with these methods differently.  

\begin{figure}
\includegraphics[width=7in]{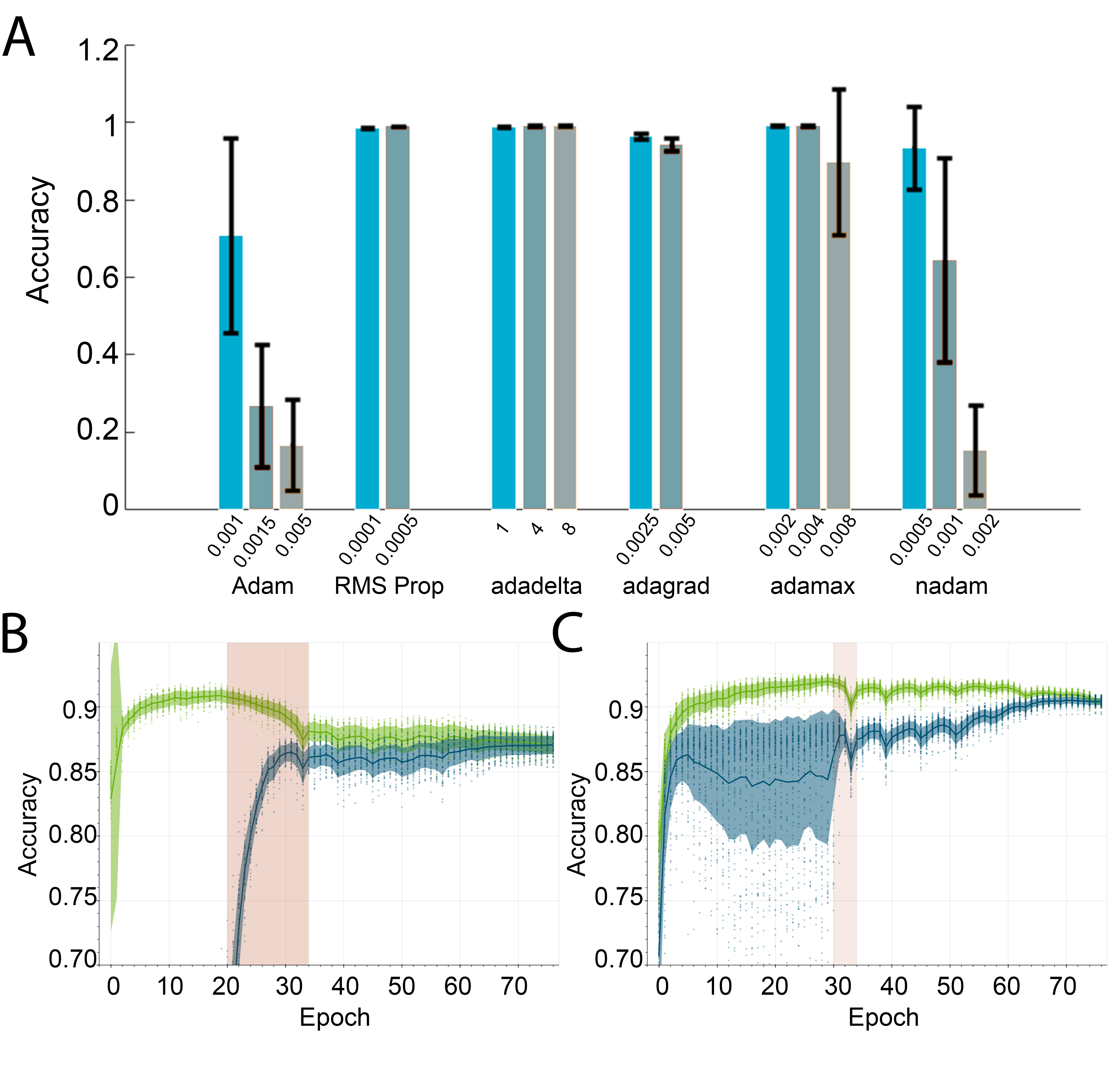}
\caption{\textbf{Network training strategy affects Whetstone spiking accuracy}.  (A) The choice of SGD optimizer has a significant impact on spiking accuracy. Some optimizers, such as Adam and nadam, suffer significant performance loss when Whetstone is used to sharpen activation functions; whereas other optimizers appear robust to the activation function perturbation introduced by Whetstone.  Error bars represent standard deviations (N=30) (B) Without batch normalization (left), severe degradation of both the test accuracy (green) and the spiking test accuracy (blue) occurs even during the sharpening of the first layer (highlighted red).  (C) The inclusion of batch normalization (right) allows for far less degradation over the same sharpening requirements. }
\label{optimizers}
\end{figure}

Fig.~\ref{optimizers}A shows the relative effect of different optimizers and learning rates on the spiking performance of Whetstone.  Often favored optimizers such as standard Adam~\cite{kingma2014adam} suffer from large performance variance.  While networks trained using Adam were at times high-performing, similar configurations would, at other times, lead to disasterous performance or total lack of convergence.  However, we did observe that some optimizers such as adadelta~\cite{zeiler2012adadelta} and adamax~\cite{kingma2014adam} can perform reliably under a variety of hyperparameters such as learning rate. This is particularly interesting as adamax is largely an $L_{\infty}$ extension of Adam; yet, with Whetstone, performance between these different optimizers differs greatly.

On experiments with MNIST and Fashion MNIST, batch normalization (BN) was found to improve the stability of accuracy by about 40 percent (Figure~\ref{optimizers}B-C). During training, BN normalizes the pre-activations of each neuron over the entire batch so that the distribution is a unit Gaussian (\ref{bneq}). This has the effect of dampening fluctuations in neuron output as parameters are modified during training, thus preventing changes in the output magnitude of earlier layers from disrupting convergence of later layers. This added stability is thought to also dampen the downstream effects of sharpening, and allows for the use of higher learning rates.
\begin{equation}
BN\left ( x_{i} \right ) = \gamma \left ( \frac{x_{i} - \mu_{B}}{\sigma_{B} + \epsilon} \right ) + \beta \label{bneq}
\end{equation}
One problem with batch normalization is that the moving averages of the normalization parameters are left in the model after training is complete. This leaves us with four extra parameters for each neuron that are used in determining pre-activations. Before we can export the model parameters to spiking hardware, is it necessary to remove these extra parameters. To accomplish this, we merge them into the weights and biases of each neuron using (\ref{bnweight}) and (\ref{bnbias}).
\begin{equation}
\text{NewWeights}\left (\mathbf{w}_{i} \right ) = \mathbf{w}_{i} \left ( \frac{\gamma}{\sigma + \epsilon} \right ) \label{bnweight}
\end{equation}
\begin{equation}
\text{NewBias}\left ( b_{i} \right ) = \left ( \frac{\gamma}{\sigma + \epsilon} \right ) \left ( b_{i} - \mu \right ) + \beta \label{bnbias}
\end{equation}

\subsection{Whetstone extends to several network types and tasks}
Finally, we looked to examine the suitability of Whetstone on ANNs designed for non-classification tasks.  As the Whetstone process is intended to be generic, we expected the process to apply to other network structures; although we expect that optimal performance will require some application-specific customization.

\begin{figure}
\includegraphics[height=7in]{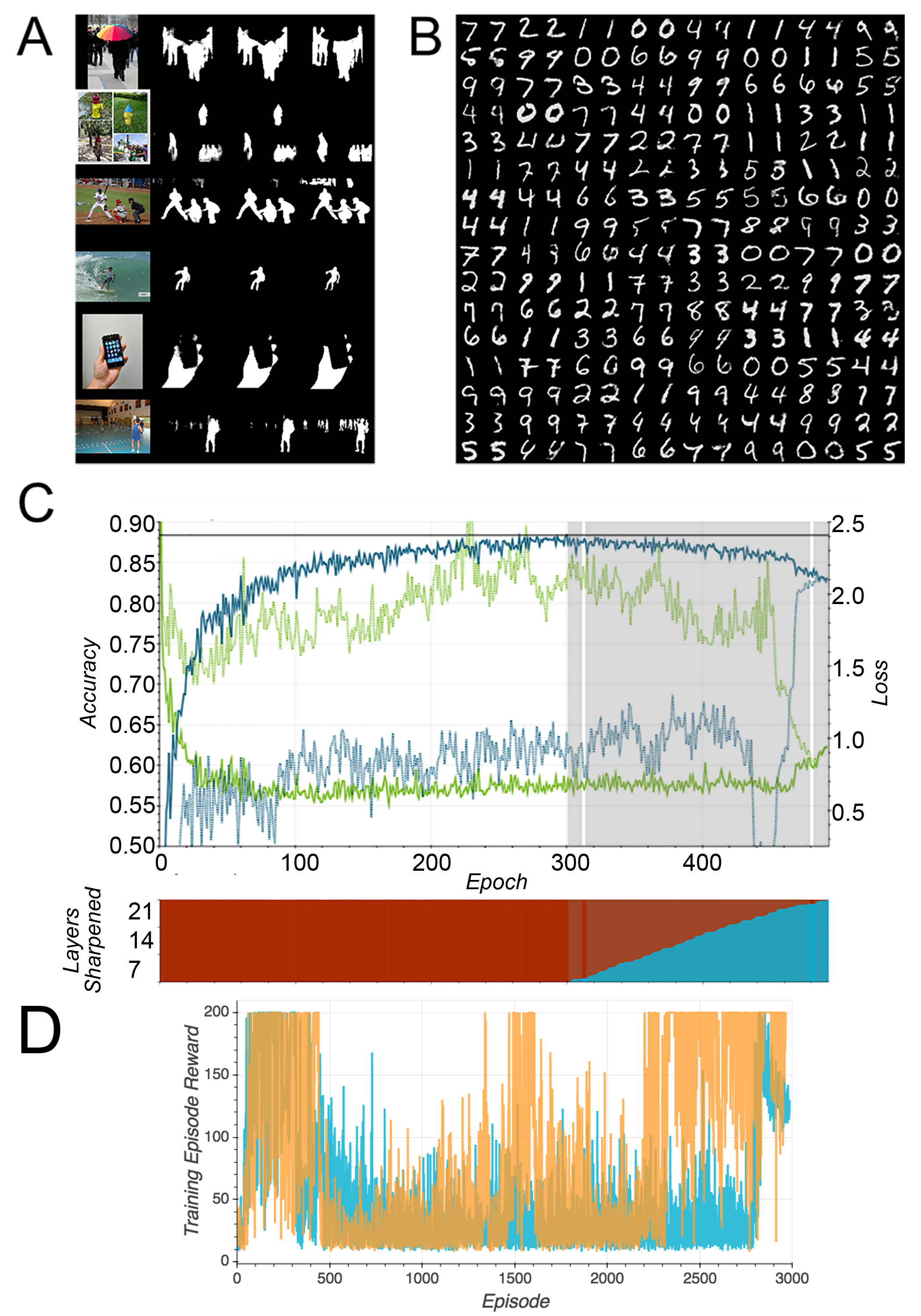}
\caption{\textbf{Whetstone has ability to sharpen non-classification networks.}  (A) ANN designed for people segmentation.  Left column illustrates original image; second column from left shows post-sharpened segmentation outputs; third column from left shows output of sharpened network with median filtering; right column shows ground truth. (B) Outputs from sharpened autoencoder are capable of reconstruction.  Left image is original image, right image is reconstructed by sharpened network. (C) Resnet architecture trained with Whetstone sharpening is capable of maintaining most of its pre-sharpening performance level. (D) Reward for each training episode for a linear (orange; final average test score $200/200$) and population code (blue; final average test score $197.92/200$) output on the CartPole reinforcement learning task.  }
\label{Alt_networks}
\end{figure}

First, we examined the performance of a 12 layer convolutional network designed to identify people in images selected from the COCO dataset \cite{lin2014microsoft}.  As shown in Fig.~\ref{Alt_networks}A, the Whetstone sharpened network was able to adequately identify people in the images, with an intersection-over-union of 0.482.  

Second, we examined the impact of Whetstone on a convolutional autoencoder designed to reconstruct MNIST images.  As shown in Fig.~\ref{Alt_networks}B, the reconstructed images are qualitatively similar to the sharpened network's inputs.  In this example, using two convolutional layers and three middle dense layers, the sharpened network could attain a binary cross-entropy of 0.647.

Next, we applied Whetstone to a Resnet architecture, which leverages non-local skip connections in performing classification (Fig.~\ref{Alt_networks}C).  While some Resnet architectures contain hundreds of intermediate modules; we tested Whetstone on a 20 layer network trained on CIFAR-10.  Prior to sharpening, this Resnet structure achieved $87.26\%$ accuracy, and after sharpening we obtained $83.11\%$ accuracy.  While this $4.15\%$ degradation is not suitable for most applications, this result was obtained without any customization of Whetstone for connections between non-sequential layers.  In particular, as shown in Fig.~\ref{Alt_networks}C, the sharpening of the last stages of the Resnet leads to much of the measured loss.  It is possible that the rather narrow network structure of Resnet may expose it to some of the challenges with `dead' bRELUs as shown in Fig.~\ref{output_encodings}.  

Of note, Resnet is an example of a network architecture with not exclusively sequential connectivity.  Resnets include ``skip connections'', whereby some layers will project to both the next layer as well as one that is several steps downstream. In conventional hardware, the retrieval of input activations is simply a memory retrieval, but in spiking neuromorphic implementations, neuron spikes are only generated at one time, requiring the use of delays between layers if there are intermediate processing stages. This use of delays provides an example of how the spiking Whetstone activations are compatible with more temporally complex processing.

Finally, we tested the ability of Whetstone to sharpen the activations of a network designed to perform deep reinforcement learning (RL) on the CartPole task.  Deep RL architectures are quite different from classic supervised learning ANNs, and as such should not immediately be assumed to be compatible with any technique for sharpening activations.  The Cartpole task is a classic reinforcement learning challenge which we believe is a strong baseline before extending to more complex tasks.  We did not seek to design a novel reinforcement learning algorithm, but rather establish the compatibility of Whetstone and existing algorithms and identify some of the challenges that exist in applying spiking networks to standard reinforcement learning taks.  

The most immediate challenge is the attribution of a continuous Q value that represents the expected reward associated with a potential decision. Q learning requires that the ability to represent many values is maintained within the network even as individual neurons become discrete in their activations.  One option is to have linear activation output neurons during training, but replace the output neurons on-hardware with a winner-take-all circuit.  Another option is to use a population code where the Q value is represented by the number of active neurons.  For this method, loss is calculated against the sum of the activations.  Another challenge is the dynamic range of the input space.  For a small task like CartPole, a small number of inputs ($4$) have a relatively large dynamic range when compared to the number of neurons typically used to solve the task.  This is generally not a problem when neurons have sufficient dynamic range and high precision, but the representational space is limited with spiking neurons.  

As with the previous examples in this section, while we did not optimize Whetstone specifically for RL, we were able to craft an experimental sharpener compatible with the RL episodes.  This enabled our dense spiking networks to `solve' the task; we trained a linear output network and a population code network with scores, averaged over $100$ testing episodes, of $197.92$ and $200$ respectively ($200$ is a perfect score).  However, these networks are extremely brittle, and convergence to an effective network is challenging, with only a small percentage of trained networks solving the task.  Figure~\ref{Alt_networks}D shows the episode reward for each episode for representative networks, and it is easy to see that the networks are highly unstable, though some of the variability is a result of purposeful exploration.  One issue is that Whetstone is (as we have seen in the classification task) best suited for wide and deep networks, whereas heavily parameterized, large networks are not appropriate for the CartPole task.  Further exploration this area of research will hopefully identify ways to better tailor Whetstone to the challenges uniquely posed by reinforcement learning.

\section{Discussion}
As recent advances in deep neural networks have yielded network architectures with more and more layers resulting in millions of parameters, various mathematical optimizations have been pursued to mitigate the associated large computational burden. These efforts include quanitization techniques reducing the precision of the weights between neurons as well as binarizing the communication. Of these optimization techniques, the BinaryConnect quantization method reduces full precision weights down to a binary -1 or 1 representation\cite{courbariaux2015binaryconnect}. BinaryNet extends this approach to also binarize the activations \cite{hubara2016binarized}. Doing so has the implication that not only is the communication analogous to spiking activity, but the computation may be implemented using the XNOR logic operation rather than requiring multiplications and additions. However, these extreme quantizations often come at a cost of impaired classification accuracy. To mathematically enable these quantizations, a set of weights for gradient training and the target reduced precision binarized weights typically must be maintained. The training process relies upon a straight-through estimator or related function to define a differentiable gradient \cite{bengio2013estimating} whereas the Whetstone method simply sharpens the activation function over time. 

Beyond mathematical optimizations a breadth of research has been done to create alternative training algorithms to backpropagation or to convert DNNs to spiking form. The former efforts include SpikeProp \cite{bohte2000spikeprop} and Gradient Descent for Spiking Neural Networks \cite{huh2017gradient}. Early work taking the latter approach include the Cao et al. approach of converting analog activations to spiking rates \cite{cao2015spiking}. Other efforts include the work of Hunsberger and Eliasmith employing a softened rate model \cite{hunsberger2015spiking,hunsberger2016training} as well as the work of Rueckauer et al. \cite{rueckauer2017conversion} providing conversions for a broad set of ANN computations such as biases and normalizations. Other approaches take into consideration implementation requirements for neuromorphic hardware such as the Energy-efficient deep neuromorphic networks (EEDN) approach which focuses upon creating convolutional neural networks whose structure is suited for IBM TrueNorth neuromorphic implementation \cite{esser15,esser2016convolutional}. Importantly, many of these techniques utilize a rate code in which multiple spikes overtime are employed to encode values. The Whetstone method conversely communicates a single bit rather than requiring a rate code. 

\begin{table}
\caption{A table summarizing results on various datasets.\label{resultTable}}
\begin{tabular}{cccc}
Algorithm/Author & Method & MNIST & CIFAR-10\\
\hline
Whetstone & Binary communication (VGG-like) & 0.9953 & 0.8467\\
Whetstone & Binary communication (10-net ensemble) & 0.9953 & 0.8801\\
Eliasmith et al. \cite{hunsberger2016training} & Transfer of trained ANN to spiking LIF   & 0.9912 & 0.8354 \\
EEDN \cite{esser15,esser2016convolutional} & TrueNorth compatible convolutional networks & 0.9942 & 0.8932 \\
Rueckauer et al. \cite{rueckauer2017conversion} & Spiking equivalents of common CNN architecture constructs & 0.9944 & 0.9085 \\
BinaryNet \cite{hubara2016binarized} & Binary weights and activations  & 0.9904 & 0.8985 \\
\end{tabular}
\label{table:comparisonTable}
\end{table}

The Whetstone method described here is intended to offer an ``off-the-shelf'' capability for machine learning practitioners to convert their deep neural network approaches to a spiking implementation suitable for neuromorphic hardware. Table \ref{table:comparisonTable} shows results for both MNIST and CIFAR-10 classification using the Whetstone method presented here as well as several of the other related techniques. While these related techniques use different network topologies, data augmentation approaches, and methodologies they share a common goal of classification performance on the presented benchmark datasets. As we show both in Fig.~\ref{performance} and Table \ref{table:comparisonTable}, with only minimal alterations, such as $N$-hot encoding, Whetstone can achieve strong performance at only a modest penalty compared to equivalent non-spiking networks.  However, some specialization, such as more extensive output encodings (Fig.~\ref{output_encodings}) and appropriate choice of the optimizer (Fig.~\ref{optimizers}) can minimize the performance cost of using spikes to communicate between neurons.

Importantly, while the method described above is well-suited for converting standard neural network techniques into a spiking form compatible with neuromorphic hardware, this approach does not yet fully take advantage of other aspects of spike-based representations that potentially offer substantial savings in power efficiency.  For instance, spiking neuromorphic platforms often leverage \textit{event-driven} communication, wherein the only information communicated are the spike-events.  Therefore, there is an energy benefit to tailor an algorithm to have sparse activities.  While Whetstone reduces the precision of communication to discrete $1$ or $0$, we currently make no attempt to sparsify the representations.  We envision that techniques leveraging sparse coding approaches could be particularly advantageous when coupled with Whetstone for this reason.  

Another aspect of SNNs is that they can encode information in the time domain.  In biological systems, \textit{when }a spike occurs often confers more information than if a spike occurs.  This temporal coding is not common to conventional ANN techniques (though it is present in some form in networks such as liquid-state machines), and since temporal coding introduces other computational trade-offs such as potentially increased latency, the value of a temporal code is limited in the applications examined here.  However, for applications such as video processing in which relevant information exists across frames, the ability for spiking neurons to integrate over time may prove useful. We see that using Whetstone to train neural networks to represent information discretely is a potential first step in a true temporal spiking code, as it preserves the temporal dynamics of neurons for use in encoding dynamic information, as opposed to relying on time to encode information that could otherwise be encoded within one time-step.  Further work is required to fully transfer neural network function into the time domain.

Not only does the Whetstone method offer a means to make use of emerging low power neuromorphic hardware, but it may be beneficial for other accelerators as well. As GPUs and and other architectures increasingly pursue the ability to perform sparse computations efficiently, the resulting binary communication from the Whetstone method is well suited for such approaches. For example, if an architecture can replace multiplications with signed addition the binarization of the communication by the Whestone method converts an ANN to a suitable representation. Or likewise a sparse multiplication which can skip multiplications by zero can make use of Whetstone networks regardless of whether the architecture is for SNNs or not.

\section{Methods}
\label{AlgorithmDescription}
\subsection{Whetstone: Converging to Spiking Activations}
\label{Methods_Whetstone}
In contrast to many methods that convert fully trained ANNs to SNNs post hoc, the Whetstone algorithm is designed to account for a target of an SNN directly into an otherwise conventional training process.  In the standard training of ANNs, for any given layer, a specific and static activation function is pre-determined.  Common activation functions include $\tanh$, sigmoid, and rectified linear units (RELUs).  In current practice, RELUs have become the standard due to quick, reliable training and high network performance.  The key insight in Whetstone is that we treat this activation function as dynamic through the training process.  In place of a static activation function, we update the activation while training progresses.  Specifically, we use a sequence of bounded, continuous functions $h_i: \mathbb{R} \rightarrow [0,1]$ such that $h_i$ approaches in measure $h$ where $h$ is the Heaviside function.  The Heaviside function is a specific parameterization of the threshold activations present on neuromorphic platforms, and each intermediate activation function is amenable to standard stochastic gradient descent methods.  We note that since neither the convergence of the weights nor of the activation functions is uniform, we have poor theoretical guarantees in most cases.  However, experimentation has shown that reliable and accurate convergence is possible in a wide variety of networks.  Additionally, in practice, we will see that it is often beneficial to leave the definition of $h_i$ for training time determination; although the core concept remains unchanged.

The convergent activation method is applicable to a variety of originating activation functions.  This implementation of Whetstone focuses on the bounded rectified linear unit (bRELU). bRELUs have been shown to be as effective or nearly as effective as RELUs, and the bounded range allows them to be easily converted to a spiking threshold function~\cite{liew2016bounded}. We parameterize our units as 
\begin{equation}
	h_{\alpha, \beta} = 
	\begin{cases}
		1,  			& \text{if } x_i \ge \beta   \\
		(x_i-\alpha)/(\beta-\alpha)       & \text{if } \alpha \le x_i < \beta   \\		
		0,  & \text{if } x_i \le \alpha,
	\end{cases}
	\label{modular_add}
\end{equation}
and assert that $|\beta - 0.5| = |\alpha - 0.5|$.  With $\alpha = 0$ and $\beta = 1$, $h_{\alpha, \beta}$ is a standard bounded RELU.  However, as $\alpha \rightarrow 0.5$, $h_{\alpha, \beta} \rightarrow h$.  After an initial period of conventional training, the spiking bRELUs are sharpened by reducing the difference between $\alpha$ and $\beta$.  The rate and method of convergence can be determined either prior to training or dynamically during training.

Fig.~\ref{OverviewWhetstone} shows the training of a standard deep convolutional network on MNIST.  As can be seen, by waiting several epochs to begin sharpening, the network can approach its eventual test accuracy. The progressive sharpening  quickly allows binarized communication networks to effectively achieve comparable performance to the non-spiking case.

\subsection{Output Encodings}
\label{Methods_Encodings}
For our classification output encoding, we use a $N$-hot representation of each class. This method has helped mitigate the fragility of the spiking networks, see Figure~\ref{output_encodings}.  Specifically, for an $N$-hot encoding, we design the networks to have the last learning layer to have N neurons for each class.  These neurons are independently initialized and have their own weights.  For determining the loss during training, there are two main options.  First, we can encode each class with its corresponding vector and use a vector-compatible loss function (e.g.~mean squared error).  Second, we can use the spiking output as a simple population code and calculate the softmax function on these embedded values.  We have found this to be the preferred method, and all classification results in this paper use the softmax method.  During training, the activation of the neurons corresponding to each class are summed, and these sums are fed into a non-learning softmax calculation.  In testing (or on hardware), we simply count the class with the most activations, which is equivalent since the softmax preserves the maximum value.  This softmax method allows us to train using crossentropy loss and still maintain compatibility with neuromorphic hardware targets.

\subsection{Sharpening Schedule}
\label{Methods_SharpeningSection}
The sharpening of networks is performed layer-by-layer, and the timing of the sharpening is determined by a schedule. Our exploration of training schedules have shown that sharpening the network from the ``bottom-up'' is more stable than from the ``top-down'' (data not shown).  This is likely due to the backwards flow of gradients during training; if top layers are sharpened first, all the nodes in the networks have reduced gradient information.

In addition to the direction of sharpening, we also examined programmed versus adaptive scheduling of sharpening.  For programmed sharpening schedules, we consider a basic paradigm where, after an initial waiting period, each layer is sharpened ``bottom-up'' over a pre-determined number of steps (either epochs or minibatches).  This method is easy to implement, but ultimately the addition of sensitive hyperparameters is undesirable. 

The adaptive sharpener is based off of a standard controls system and uses the training loss to automatically decide when to start and stop sharpening, freeing the user from having to craft a sharpening schedule manually. At the end of each epoch, it looks at the change in training loss, and uses it to decide whether to sharpen for the next epoch, or pause for several epochs. When in a sharpening state, if the loss increases by more than a specified percentage, then sharpening is halted. When in a non-sharpening state, if the loss fails to improve more than a certain percentage after a certain number of epochs, then sharpening resumes. The sharpening rate is specified as the amount per layer per epoch, where amount is a floating point value less than or equal to $1.0$. For example, if the sharpening rate is set to $0.25$, then it will take four epochs in the sharpening state to completely sharpen one layer. It is important to note that the sharpness of a layer is altered at the end of each batch, providing a more gradual transition than if it were altered at the end of each epoch. Our experience suggests that frequent, small updates are beneficial.  This process is outlined as a state diagram in Fig.~\ref{WhetstoneMethod}. In this state diagram, transition rules are only evaluated at the end of each training epoch. `Wait' states halt sharpening for one epoch of training. Depending on the sharpening mode, the `Sharp' state will either sharpen all model layers (for uniform) or just the current layer (for bottom-up). The process terminates when all layers of the model have been fully sharpened.

\subsection{Implementation and Software Package Details}
\label{Methods_Implementation}
Whetstone is a hybrid method, intended to provide users with a familiar interface while enabling translation to specialized hardware.  Our implementation is thus intended to be compatible with conventional deep learning tools at the software level, while providing a network output suitable for implementation on spiking neuromorphic hardware (or other specialized hardware that can benefit from discrete communication). Whetstone is implemented as a set of custom Keras-compatible modules~\cite{chollet2015keras}.  We have performed extensive testing using the Tensorflow backend though as Whetstone is pure Keras, it should automatically support all underlying backends such as Theano and CNTK.  

Because of the challenges associated with spiking algorithms, the implementation of Whetstone was designed with the goal to `speak the language of the DL researcher,' so as to minimize the burden on the user. Applied here, this principle means that specifics of the underlying spiking neural network should be abstracted away and that there should be a minimal disruption to the workflow.  Compared to a standard Keras model, Whetstone-ready models generally have three modifications:
\begin{itemize}
\item Spiking Activations:  Standard RELU or sigmoid activations need to be replaced with the parameterized spiking versions provided by the Whetstone library.
\item Sharpening Callback: A sharpening callback must be attached during the training process.  This can be simplified using a standard dynamic, adaptive sharpener or by hand-selecting stepping points (see~\ref{Methods_SharpeningSection}).
\item Output Encoding:  For classification problems, it is standard practice for a network to compute a softmax activation on its logits.  However, spiking platforms do not innately support this function.  Instead, we wrap an output layer (with possible redundant population encoding, see~\ref{EncodingSection}) in a non-learning softmax layer which decodes any population code.  On hardware, this layer can either be computed on a host machine or the raw number of spikes can be used in a winner-take-all circuit.
\end{itemize}

The network is then trained as usual, using any methods and packages compatible with Keras (e.g. hyperas, opencv, etc.). Once training is completed, the final Keras model can be directly transferred to a leaky-integrate-and-fire neuron model which are compatible with spiking neural hardware.  The resulting networks can be simulated either on CPUs/GPUs or implemented on neuromorphic hardware using a tool such as N2A or PyNN~\cite{rothganger2014n2a,davison2009pynn}.  

\subsubsection{Data Availability Statement}
All data used comes from publicly available datasets: MNIST~\cite{lecun2010mnist}, Fashion-MNIST~\cite{xiao2017fashion}, CIFAR~\cite{krizhevsky2009learning}, and COCO~\cite{lin2014microsoft}.

\subsubsection{Code Availability Statement}
Whetstone is available at \url{https://github.com/SNL-NERL/Whetstone}, licensed under the GPL.

\subsection*{Acknowledgments}
This work was supported by Sandia National Laboratories' Laboratory Directed Research and Development (LDRD) Program under the Hardware Acceleration of Adaptive Neural Algorithms Grand Challenge project and the DOE Advanced Simulation and Computing program.  Sandia National Laboratories is a multi-mission laboratory managed and operated by National Technology and Engineering Solutions of Sandia, a wholly owned subsidiary of Honeywell International, Inc., for the U. S. Department of Energy's National Nuclear Security Administration under Contract DE-NA0003525.  

This paper describes objective technical results and analysis. Any subjective views or opinions that might be expressed in the paper do not necessarily represent the views of the U.S. Department of Energy or the United States Government.

\end{document}